\documentclass{article}

\usepackage{arxiv}

\usepackage[utf8]{inputenc} 
\usepackage[T1]{fontenc}    
\usepackage{hyperref}       
\usepackage{url}            
\usepackage{booktabs}       
\usepackage{amsfonts}       
\usepackage{nicefrac}       
\usepackage{microtype}      
\usepackage{lipsum}
\usepackage{graphicx}
\usepackage[sort,numbers]{natbib}
\usepackage{mathrsfs}
\usepackage{array}
\usepackage[title]{appendix}
\counterwithin{table}{section}
\counterwithin{figure}{section}
\graphicspath{ {./Figures/} }
\usepackage{fancyhdr}
\title{SKIPP'D: a \textbf{SK}y \textbf{I}mages and \textbf{P}hotovoltaic \textbf{P}ower Generation \textbf{D}ataset for Short-term Solar Forecasting}

\author{
 Yuhao Nie\footnotemark[1]\\
 Department of Energy Resources Engineering\\
 Stanford University \\
 Stanford, CA 94305, USA\\
 \And
 Xiatong Li\thanks{The first two authors have equal contribution.}\\
  Department of Civil and Environmental Engineering\\
  Stanford University \\
  Stanford, CA 94305, USA\\
 \And
 Andea Scott\\
 Department of Energy Resources Engineering\\
 Stanford University\\
 Stanford, CA 94305, USA\\
 \And
 Yuchi Sun\thanks{Current affiliation: Energy and Environmental Economics, Inc., San Francisco, CA 94104, USA}\\
 Department of Energy Resources Engineering\\
 Stanford University\\
 Stanford, CA 94305, USA\\
 \And
 Vignesh Venugopal\footnotemark[2]\\
 Department of Energy Resources Engineering\\
 Stanford University\\
 Stanford, CA 94305, USA\\
 \And
 Adam Brandt\thanks{Corresponding author: \texttt{abrandt@stanford.edu}}\\
 Department of Energy Resources Engineering\\
 Stanford University\\
 Stanford, CA 94305, USA\\
  }

\let\oldmaketitle\maketitle
\renewcommand{\maketitle}{\oldmaketitle\setcounter{footnote}{0}}

\begin{document}
Highlights
\begin{enumerate}
    \item A curated sky images and photovoltaic power generation dataset (SKIPP’D) is introduced to facilitate the research and benchmark of image-based solar forecasting.
    \item Three years of high temporal frequency processed data and overlapping high resolution raw data are both provided for flexibility of research.
    \item A code base containing data processing scripts and baseline model implementations is provided.
    \item Baseline deep learning models are developed for two different solar forecasting tasks to demonstrate the usage of the dataset.

\end{enumerate}
\newpage
\maketitle
\thispagestyle{fancy}
\begin{abstract}
Large-scale integration of photovoltaics (PV) into electricity grids is challenged by the intermittent nature of solar power. Sky-image-based solar forecasting using deep learning has been recognized as a promising approach to predicting the short-term fluctuations. However, there are few publicly available standardized benchmark datasets for image-based solar forecasting, which limits the comparison of different forecasting models and the exploration of forecasting methods. To fill these gaps, we introduce SKIPP'D --- a \textbf{SK}y \textbf{I}mages and \textbf{P}hotovoltaic \textbf{P}ower Generation \textbf{D}ataset. The dataset contains three years (2017-2019) of quality-controlled down-sampled sky images and PV power generation data that is ready-to-use for short-term solar forecasting using deep learning. In addition, to support the flexibility in research, we provide the high resolution, high frequency sky images and PV power generation data as well as the concurrent sky video footage. We also include a code base containing data processing scripts and baseline model implementations for researchers to reproduce our previous work and accelerate their research in solar forecasting.
\end{abstract}


\section{Introduction}
Solar PV is rapidly becoming a significant source of power generation. Fluctuations in solar power generation due to short-term events (like moving clouds) can have large impacts in areas with high solar PV penetration. Images captured by ground-based fish-eye cameras contain a wealth of information about the sky, but this information is challenging to extract and use for reliable predictions. In the past five years, using emerging deep learning models to “read” the sky and make forecasts of PV power generation (or solar irradiance) has shown promising performance \cite{Sun2018,Zhang2018,Sun2019,Nie2020,Feng2020,Paletta2021,Paletta2021eclipse,Feng2022}.

However, two major challenges have been identified in this fast growing area. First, prior work is hard to compare as the models are developed using different datasets with different specifications. There is a lack of standardized datasets for benchmarking deep-learning-based solar forecasting models. Secondly, deep learning models are data hungry. To make deep learning models generalize well, it often requires massive and diversified training data. Several different parties have been contributing to addressing the data accessibility issues. The first type of party is comprised of national labs and research organizations which have multi-year efforts in data collection and publication of archived datasets to the public. For example, Solar Radiation Research Laboratory (SRRL) \cite{stoffel1981nrel} from NREL developed a Baseline Measurement System (BMS), which open sources multi-years of 10-min 
high resolution sky images from two different sky cameras together with 1-min radiation and meteorological measurements in Golden, Colorado. However, the low temporal resolution of the image data can hardly satisfy the need of short-term solar forecasting due to the variation and volatility of the clouds. Another example is the National Surface Radiation (SURFRAD) Budget Network developed by National Oceanic and Atmospheric Administration (NOAA) \cite{augustine2000surfrad}. It provides an archived dataset of sky images and radiation and meteorological sensor measurement data from multiple sites across United States for public use, but the sky images are usually logged in low-resolution (e.g., 288$\times$352) and low temporal frequency (e.g., 1 hour). The second type of party is research groups sharing the solar forecasting datasets used in their publications. One of the most comprehensive datasets for solar forecasting was compiled by \citet{Pedro2019}, which includes 3-years of high-resolution sky images (1536$\times$1536) and irradiance measurements in 1-min frequency collected at Folsom, California, together with overlapping data from satellite imagery and Numerical Weather Prediction forecasts, as well as the features extracted from the imagery and irradiance data. Other efforts include \textit{SkyCam} by \citet{Ntavelis2021} which contains 1-year of sky images (600$\times$600) collected from three different locations in Switzerland together with the overlapping irradiance measurements both logged in 10-second frequency, and \textit{Girasol} by \citet{terren2021girasol} which provides high temporal frequency sky images from both visible and infrared cameras and irradiance measurements data for 244 days, while the visible images only have one intensity channel and the infrared images are of low resolution. The datasets released by \citet{dissawa2021sky} and \citet{bassous2021development} both provide sky images and PV power output data, but the sizes are pretty small as they only include a few days/months of data. Finally, a third type of party is researchers developing open-source tools for easy access of publicly available datasets, most of which are from national labs or research organizations as described previously. Such efforts include \citet{yang2018solardata}'s \textit{SolarData} and \citet{Feng2019Opensolar}'s \textit{OpenSolar}. 

Despite the various efforts we mentioned above, there are limited publicly available datasets that are compiled with high quality data and are suitable for deep-learning-based or computer-vision-based short-term solar forecasting research. To fill these gaps, we introduce SKIPP'D --- a \textbf{SK}y \textbf{I}mages and \textbf{P}hotovoltaic \textbf{P}ower Generation \textbf{D}ataset, collected and compiled by the Environmental Assessment and Optimization (EAO) Group at Stanford University. The dataset contains the following two levels of data which  distinguishes it from most of the existing open-sourced datasets and makes it especially suitable for deep-learning-based solar forecasting research:
\begin{enumerate}
    \item Benchmark dataset: 3 years of processed sky images (64$\times$64) and concurrent PV power generation data with a 1-min interval that are ready-to-use for deep learning model development;
    \item Raw dataset: Overlapping high resolution sky video footage (2048$\times$2048) recorded at 20 frames per second, and sky image frames (2048$\times$2048) and history PV power generation data logged with a 1-min frequency that suit various research purposes.
\end{enumerate}
In addition, we provide a code base containing data processing scripts and
baseline model implementations for researchers to quickly reproduce our previous work and accelerate solar forecasting research. We hope that this dataset will facilitate the research of image-based solar forecasting and contribute to a standardized benchmark for evaluating and comparing different solar forecasting models. Besides, we also encourage the users to explore on solar forecasting related areas with this dataset, such as sky image segmentation and cloud movement forecasting.

\section{Data Sources}
\label{sec:data_sources}
Our research group started the data collection from March 2017 at Stanford University's campus, located in the center of the San Francisco Peninsula, in California. According to the Köppen climate classification system, Stanford  has a warm-summer Mediterranean climate, abbreviated Csb (C=temperate climate s=dry summer b=warm summer) on climate maps \cite{kottek2006}. In terms of cloud coverage, Stanford is featured by long summers with mostly clear skies and short winters with partly cloudy skies. Two major categories of data are collected and logged: sky images and PV power generation, which are detailed in Section \ref{subsec:sky_images} and \ref{subsec:pv_power_generation}, respectively. Over the past five years, our lab has collected over two terabytes of data, which have enabled multiple published solar forecasting studies \cite{Sun2018,Sun2019,Sun2019_dissertation,Venugopal2019,Venugopal2019thesis,Nie2020,Nie2021}.

In this release, we open-source the data from March 2017 to December 2019 \footnote{The dataset suffers from some interruptions due to the water intrusion, wiring and/or electrical failure of the camera, as well as daylight-saving adjustment failure of the camera in 2017 and 2018, which is back to normal for 2019 and beyond.}. Here, we provide two levels of data to suit the different needs of researchers: (1) A processed dataset consists of 1-min down-sampled sky images (64$\times$64) and PV power generation pairs, which is intended for fast reproducing our previous work and accelerating the development and benchmarking of deep-learning-based solar forecasting models; (2) A raw dataset consisting of high resolution sky images (2048$\times$2048) and PV power generation data, as well as the source sky video footage, which is intended for customizing data extraction and exploring other related areas of solar forecasting such as cloud segmentation and cloud movement forecasting. The specifications of the data are summarized in Table \ref{table:specifications_table} and a full description of the data files can be found in Appendix \ref{appendix:data_repo}.

\begin{table}[h!]
\begin{center}
\begin{small}
\caption{Specifications Table}
\label{table:specifications_table}
\begin{tabular}{>{\raggedright}p{0.25\linewidth}>{\raggedright\arraybackslash}p{0.7\linewidth}}
\hline \hline
\noalign{\vskip 1mm}
Subject area & Solar forecasting; Computer vision; Deep learning\\
\noalign{\vskip 1mm}
More specific subject area & PV power generation prediction; Sun tracking; Cloud detection; Cloud movement prediction \\
\noalign{\vskip 1mm}
Data collection period & March 2017 to December 2019 \\
\noalign{\vskip 1mm}
Type of data &  1. Processed (benchmark) data: 1-min 64$\times$64 sky images (\textit{.npy}) and PV power generation (\textit{.npy}) pairs, partitioned into model development set (training+validation) and test set, and further structured and stored as \textit{.hdf5} format.\\
\noalign{\vskip 1mm}
 &  2. Raw data: 2048$\times$2048 sky videos recorded at 20 frames per second (\textit{.mp4}), 1-min 2048$\times$2048 sky images (\textit{.jpg}) and PV power generation (\textit{.csv}) \\
\noalign{\vskip 1mm}
How data was acquired & The sky videos/images are acquired by the camera installed on top of the Green Earth Sciences Building (37.427$^{\circ}$, -122.174$^{\circ}$) at Stanford University. The PV power generation data are from PV panel approximately 125 meters away from the camera on the roof of the Jen-Hsun Huang Engineering Center at Stanford University, which are logged by Stanford Utility and shared to us. \\
\noalign{\vskip 1mm}
Data accessibility & The processed (benchmark) data is available at \url{https://purl.stanford.edu/dj417rh1007} and the raw data is deposit separately by each year given its large size. The 2017 raw data is available at \url{https://purl.stanford.edu/sm043zf7254} and the links to 2018 and 2019 data can be found in the "Related items" elsewhere on the same web page. \\
\noalign{\vskip 1mm}
GitHub repository & Code base of data processing and baseline model are available at the GitHub repository \url{https://github.com/yuhao-nie/Stanford-solar-forecasting-dataset}.\\
\noalign{\vskip 1mm}
\hline \hline
\end{tabular}
\end{small}
\end{center}
\end{table}

In future releases, we will open source the data from 2020 and beyond of the Stanford dataset and include two additional data sources \footnote{The updated information will be released on our dataset GitHub repository: \url{https://github.com/yuhao-nie/Stanford-solar-forecasting-dataset}}: sky images and PV power generation data from a solar farm in Oregon collected by our research group and sky images from cameras set up by NREL which correspond to solar irradiance data collected by them. The solar farm in Oregon is a 3.08 MW-DC ground mounted solar array located outside of Sheridan, a rural community. Sheridan is located in the Willamette Valley which, like Stanford, has a warm-summer Mediterranean climate. However, unlike Stanford, the winters are long and typically characterized by cloudy skies and periods of rain. Our team set up a fish-eye camera (Hikvision DS-2CD6365GOE-IVS) at this location at the end of 2021 to start collecting data. In addition to sky images from Oregon, our group started collecting sky images in 2021 from NREL's Solar Radiation Research Laboratory in Golden, Colorado via their website \footnote{ NREL Solar Radiation Laboratory Baseline Management System website: \url{https://midcdmz.nrel.gov/apps/sitehome.pl?site=BMS}}, which provides a live view of the sky image every 60 seconds. Unfortunately, NREL is currently only storing images every 10 minutes, so our team is collecting and storing the minutely images in order to have a dataset that is useful to the short-term solar forecasting community. These images can be paired with NREL's minutely irradiance data which is available via their website.

\subsection{Sky Images}
\label{subsec:sky_images}
\begin{figure}[h!]
\includegraphics[width=1.0\textwidth]{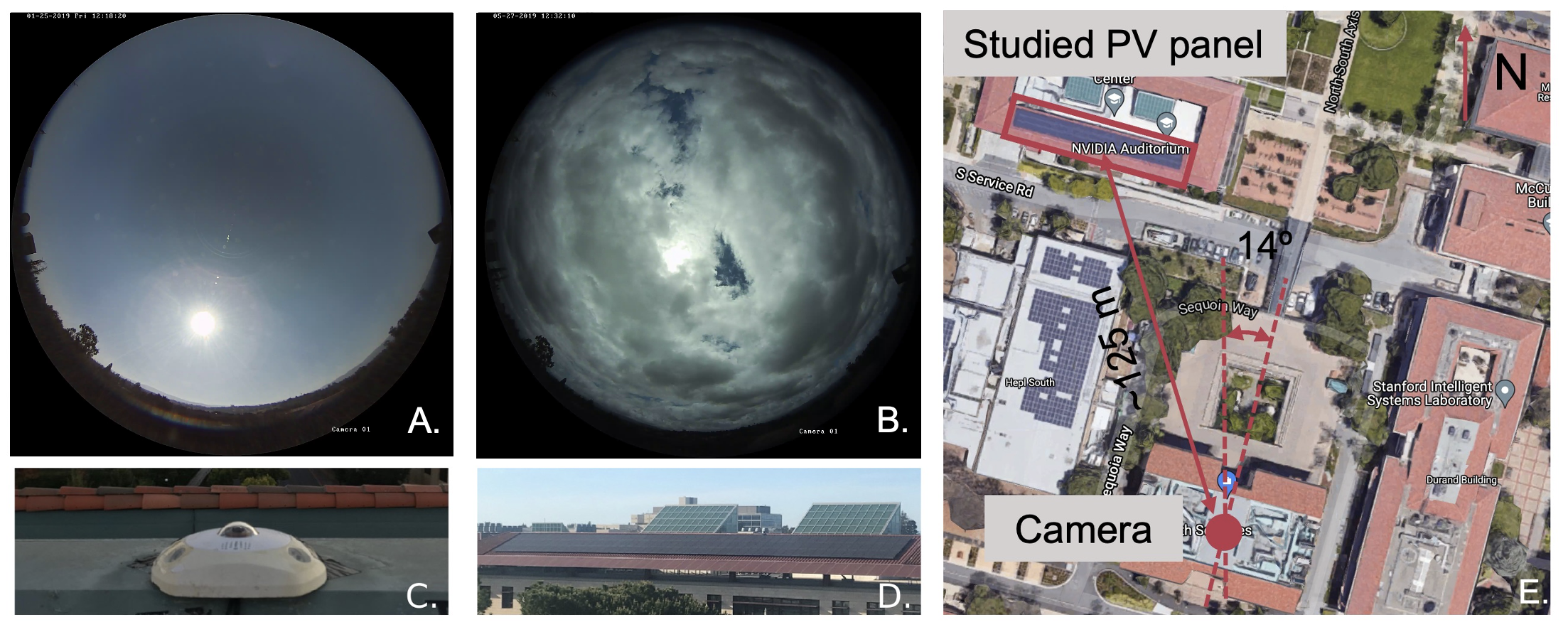}
\caption{Photos of Sky Images and Research Equipment (A. Sky image captured on a clear day at 12:18:20 pm, January 25, 2019. B. Sky image of a cloudy day captured at 12:32:10 pm May 27, 2019. C. Fish-eye camera used for sky imaging. D. Studied PV panels. E. Locations of the camera and studied solar panels).  Adapted from \citet{Sun2018}, used with permission.}
\label{fig:equipment_photo}
\end{figure}

The sky images are frames from videos recorded during daytime (6:00 AM $\sim$ 8:00 PM \footnote{Data were recorded based on the local time zone, which is either Pacific Standard Time (PST) or Pacific Daylight Time (PDT). In US, PDT starts on the second Sunday in March and ends on the first Sunday in November. }) by a 6-megapixel 360-degree fish-eye camera (Hikvision DS-2CD6362F-IV2\footnote{The camera model Hikvision DS-2CD6362F-IV is discontinued and is replaced by a new model Hikvision DS-2CD6365GOE-IVS. We replace the old model with the new model on April 29, 2022 due to aging.}) located on top of the Green Earth Sciences Building (37.427$^{\circ}$, -122.174$^{\circ}$) at Stanford University and oriented towards 14° south by west. Compared with other high-end commercial sky imaging systems, this off-the-shelf network camera is more accessible and affordable for sky monitoring. The camera holds constant camera aperture, white balance and dynamic range and captures video with a resolution of 2048$\times$2048 pixels at 20 frames per second (fps). The images (\textit{.jpg}) are extracted from the video at 1-min sampling frequency and are down-sampled to a resolution of 64$\times$64 pixels, which is found to be acceptable for PV output forecast while retaining reasonable model training time \cite{Sun2018}. Figure \ref{fig:equipment_photo} gives examples of sky images in different weather conditions, and shows the locations of the camera and PV panels used in this collection.

\subsection{PV Power Generation}
\label{subsec:pv_power_generation}
The PV output data are collected from solar panel arrays approximately 125 meters away from the camera, situated on the top of the Jen-Hsun Huang Engineering Center at Stanford University, with and elevation angle of 22.5$^{\circ}$ and an azimuth angle of 195$^{\circ}$. The PV panels are manufactured with poly-crystalline technology and the system is rated at 30.1 kW-DC. The PV output generation data are logged with 1-min frequency and are minutely averaged. The forward average is applied, e.g., value at 8:00:00 am representing the average PV generation from 8:00:00 to 8:00:59 am.

\section{Data Processing}
\label{sec:data_processing}
To support the flexibility of research, we also open source high-resolution, high-frequency raw data, and the users can customize their own data processing pipeline and process the data based on their needs. Here, we provide a reference by going over the data processing steps we used to generate the processed (benchmark) data. The processing steps were largely used in our previous published work except some minor modifications \footnote{This dataset have provided the minutely average raw PV data, so users do not need to take rolling average during data processing to get the minutely average data. This is the case in previous published work \cite{Sun2018,Sun2019} as we used the instantaneous raw PV data.}. We also provide the code base of data processing for users' reference and the descriptions of the code files can be found in Appendix \ref{appendix:sample_code}.

The data processing basically includes the following four major steps. More details on the data processing steps can be found in the PhD dissertation by \citet{Sun2019_dissertation}.

\begin{enumerate}
    \item Obtain raw high-resolution image frames (2048$\times$2048) by snapshotting the video footage at a designated frequency. While 1-min sampling frequency was used in the dataset, it is freely adjustable.
    \item Process raw PV power generation history, which includes the following two sub-steps:
    
    \begin{enumerate}
         \item Interpolate PV data to every 10 seconds in preparation for matching with images in Step 3 with irregular time stamps, e.g., 08:20:10 (raw PV data are logged regularly as 08:00:00, 08:01:00, 08:02:00, etc.)
        \item Filter out the PV data that are abnormally recorded (e.g., data logger repeatedly logs one value for a certain time period), negative (e.g., night time) or have missing records larger than 1 hour.
    \end{enumerate}
    
    \item Process raw high resolution images and pair the processed images with the concurrent processed PV generation data. The image processing includes the following two sub-steps:
    \begin{enumerate}
        \item Down-sample the high resolution image frames into low resolution (64$\times$64) for sake of saving training time. Down-sampling rate is adjustable.
        \item Filter out erroneously repeating images caused by the occasional abnormal behavior of OpenCV video decoder FFmpeg.
    \end{enumerate}
\end{enumerate}
We further partition the processed dataset into development set (including data for training and validating the model) and test set. The test set includes 10 sunny days and 10 cloudy days selected manually across 2017 to 2019; the rest of the data goes into the development set. The number of samples contained in the model development set and the test set are 349,372 and 14,003, respectively, roughly a 96\%:4\% split. It should be noted that we have updated the test days from our previous publications given that missing data points were observed in certain days of the old test set. For comparison with the results from our previous publications, we encourage the users to re-organize the data by themselves. We hope this processed dataset can be used as a standardized benchmark for model training, evaluating and comparison in the solar forecasting community.

\section{Benchmark Dataset for Image-based Solar Forecasting Using Deep Learning}

\begin{figure}[h!]
\includegraphics[width=1.0\textwidth]{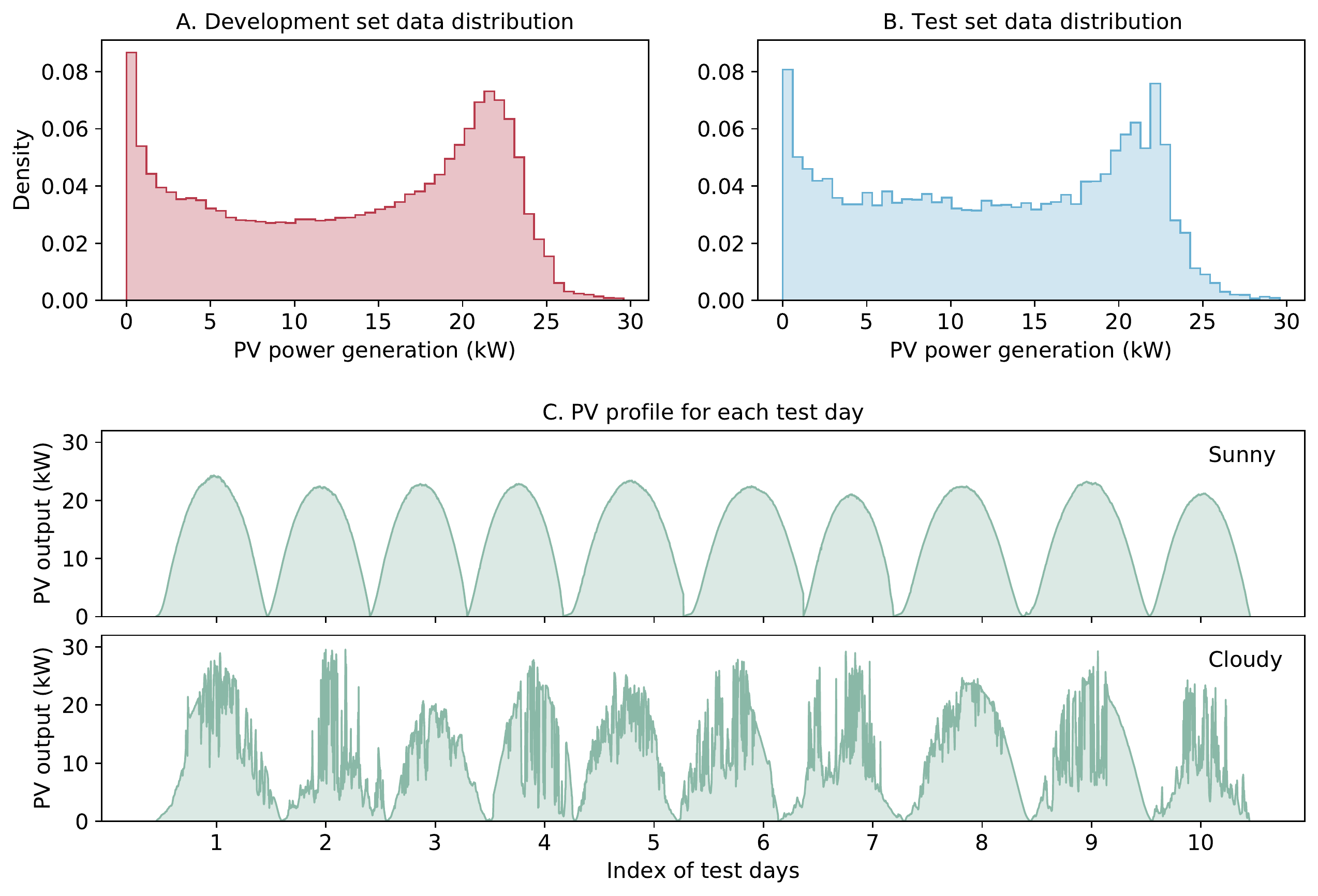}
\caption{The PV power generation data distribution of the benchmark dataset: A. development set PV data distribution; B. test set PV data distribution; and C. the PV power generation profiles of the 10 sunny days and 10 cloudy days used in the test set: upper panel shows the sunny days, and the lower panel is for the cloudy days.}
\label{fig:combined_pv_distribution_test_set_pv_profiles}
\end{figure}

The benchmark dataset ($\mathscr{D}$) contains the model development set ($\mathscr{D}_d$) and test set ($\mathscr{D}_t$) obtained from the data processing steps described in \ref{sec:data_processing}. The samples of the benchmark dataset are organized as aligned pairs of sky images ($\mathcal{I}$) and PV power generation ($\mathcal{P}$), i.e., $\mathscr{D} = \{(\mathcal{I}_i, \mathcal{P}_i) \mid i\in \mathbb{Z}: 1\leq i\leq N\}$, where $N=363,375$ is the total number of samples in the benchmark dataset.  Figure \ref{fig:combined_pv_distribution_test_set_pv_profiles} shows the distribution of the PV power generation data for the development set and test set and the PV power generation profiles of the 20 days in the test set. The statistics of the 20 test days are listed in Table \ref{table:list_testdays}.

%

\begin{table}[h!]
\begin{center}
\begin{small}
\caption{Statistics of the 10 sunny and 10 cloudy days used in the test set}
\label{table:list_testdays}
\begin{tabular}{p{0.1\linewidth}p{0.1\linewidth}p{0.1\linewidth}p{0.1\linewidth}p{0.1\linewidth}}
\hline \hline
\noalign{\vskip 1mm}
Date & Index & Mean (kW) & Max (kW) & Std (kW) \\
\noalign{\vskip 1mm}
\hline
\noalign{\vskip 1mm}
2017-09-15 &   Sunny\_1 & 15.23 & 24.35 & 7.83 \\
\noalign{\vskip 1mm}
2017-10-06 &   Sunny\_2 & 14.65 & 22.44 & 7.03 \\
\noalign{\vskip 1mm}
2017-10-22 &   Sunny\_3 & 15.17 & 22.86 & 6.97 \\
\noalign{\vskip 1mm}
2018-02-16 &   Sunny\_4 & 15.28 & 22.89 & 6.86 \\
\noalign{\vskip 1mm}
2018-06-12 &   Sunny\_5 & 14.69 & 23.45 & 7.60 \\
\noalign{\vskip 1mm}
2018-06-23 &   Sunny\_6 & 14.30 & 22.48 & 7.32 \\
\noalign{\vskip 1mm}
2019-01-25 &   Sunny\_7 & 14.08 & 21.07 & 6.26 \\
\noalign{\vskip 1mm}
2019-06-23 &   Sunny\_8 & 13.31 & 22.45 & 7.80 \\
\noalign{\vskip 1mm}
2019-07-14 &   Sunny\_9 & 13.72 & 23.27 & 7.98 \\
\noalign{\vskip 1mm}
2019-10-14 &  Sunny\_10 & 13.56 & 21.22 & 6.67 \\
\noalign{\vskip 1mm} \hline \noalign{\vskip 1mm}
2017-06-24 &  Cloudy\_1 & 12.39 & 28.95 & 8.30 \\
\noalign{\vskip 1mm}
2017-09-20 &  Cloudy\_2 & 7.55  & 29.57 & 6.88 \\
\noalign{\vskip 1mm}
2017-10-11 &  Cloudy\_3 & 10.64 & 20.72 & 6.04 \\
\noalign{\vskip 1mm}
2018-01-25 &  Cloudy\_4 & 12.39 & 27.77 & 8.08 \\
\noalign{\vskip 1mm}
2018-03-09 &  Cloudy\_5 & 12.45 & 25.60 & 7.07 \\
\noalign{\vskip 1mm}
2018-10-04 &  Cloudy\_6 & 11.83 & 27.82 & 6.76 \\
\noalign{\vskip 1mm}
2019-05-27 &  Cloudy\_7 &  8.62 & 29.22 & 7.39 \\
\noalign{\vskip 1mm}
2019-06-28 &  Cloudy\_8 & 13.25 & 24.70 & 7.68 \\
\noalign{\vskip 1mm}
2019-08-10 &  Cloudy\_9 & 11.14 & 29.28 & 7.48 \\
\noalign{\vskip 1mm}
2019-10-19 & Cloudy\_10 & 7.71 & 24.28  & 6.19 \\
\noalign{\vskip 1mm}
\hline \hline
\end{tabular}
\end{small}
\end{center}
\end{table}

\section{Sample Uses of the Benchmark Dataset}
\begin{figure}[h!]
\includegraphics[width=1.0\textwidth]{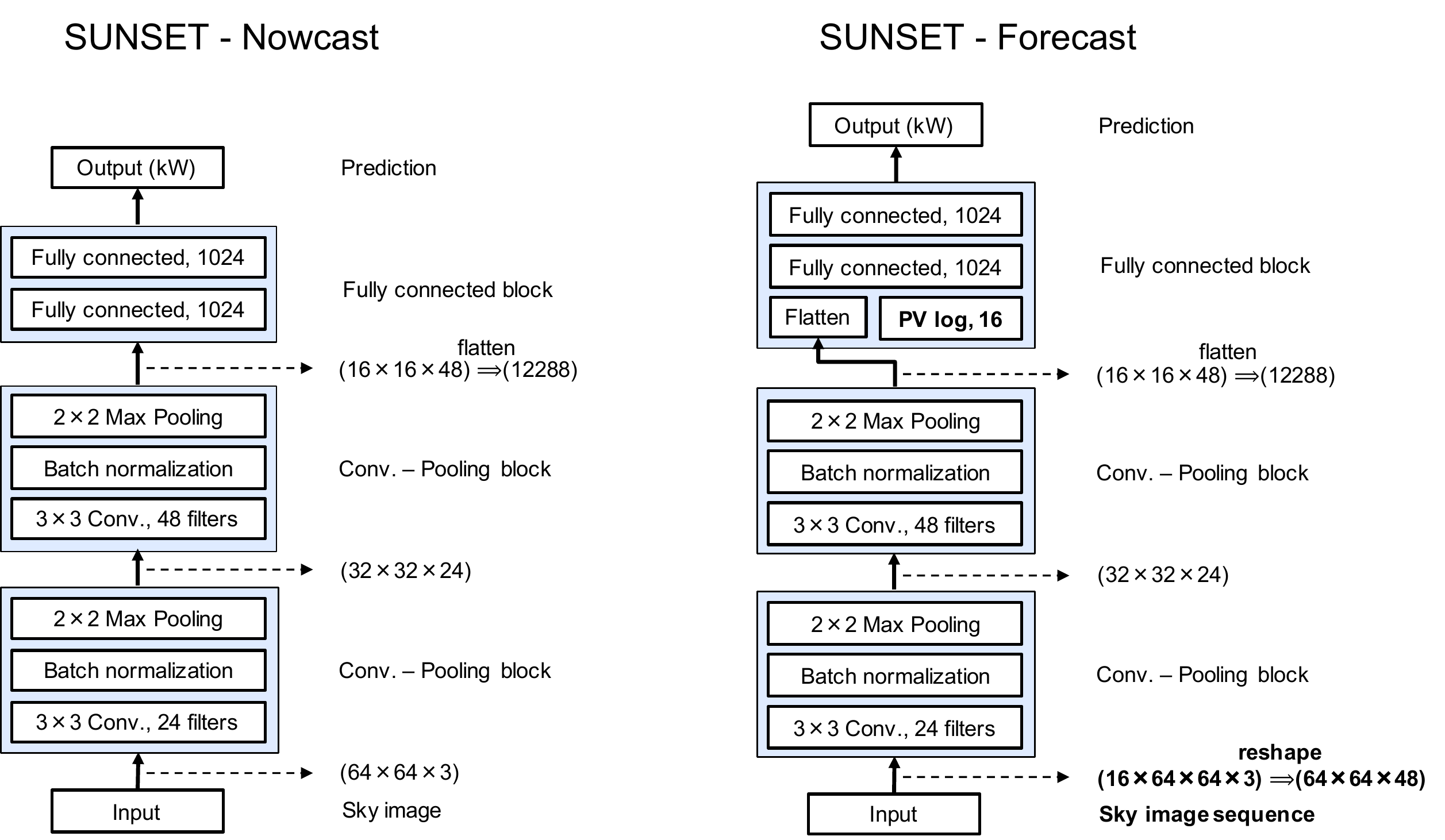}
\caption{The model architectures for PV power generation nowcast and forecast (differences between the two model architectures are highlighted in bold font). Adapted from \citet{Nie2021}, used with permission.}
\label{fig:model_arch}
\end{figure}

In this section, we demonstrate two use cases of the benchmark dataset based on our published works. Our group has developed a specialized convolutional neural network (CNN) model named SUNSET (Stanford University Neural Network for Solar Electricity Trend) for PV power generation prediction. Two specific prediction tasks were investigated based on SUNSET, including (1) PV power generation nowcast \cite{Sun2018}, i.e., given a sky image, predicting the contemporaneous PV output; and (2) PV power generation forecast \cite{Sun2019}, given sky images and PV output for the past 15 minutes on 1-minute resolution, predicting PV output 15 minutes ahead into the future. Figure \ref{fig:model_arch} shows the model architectures for the nowcast and forecast tasks. The details of these two models can be found in the corresponding published papers, and we demonstrate the use cases based on the setups described in these papers. We implemented these two deep learning models using TensorFlow 2.x, which is an update from the TensorFlow 1.x code base used in our previous publications. The new code base is included along with the dataset as described in Appendix \ref{appendix:sample_code}.

The performance of each model is evaluated by the following metrics: mean absolute error (MAE), root mean squared error (RMSE), and additionally, forecast skill (FS) for evaluating the SUNSET forecast model, which is computed based on the RMSE and is relative to the persistence model. The detailed method for calculating the FS can be found in \cite{Sun2019}. The codes for model evaluation are included in the model implementation codes (see Appendix \ref{appendix:sample_code}). It should be noted that the goal here is to demonstrate the value of the data by providing some sample use cases, rather than evaluating specific methodologies. Therefore, our analysis on the sample results will mostly be high-level.

\subsection{PV Power Generation Nowcast}
The PV power generation nowcast task is essentially learning a mapping ($f_N$) from sky images to the simultaneous PV power generation. An analog one can think of is the computer vision task estimating the age of people based on their facial images.
\begin{equation}
   f_N:\mathcal{I}_i\mapsto \mathcal{P}_i, \mathrm{where}\ \{\mathcal{I}_i, \mathcal{P}_i\} \in \mathscr{D}
\end{equation}
We developed the nowcast model based on $\mathscr{D}_d$ and evaluated the model based on $\mathscr{D}_t$. During the model development phase, ten-fold cross-validation is employed, and during the test phase, the prediction is represented by the ensemble mean of the predictions from the 10 submodels. Table \ref{tab:performance_metrics} shows the performance of the SUNSET nowcast model on the test set evaluated by RMSE and MAE. Figure \ref{fig:sunset_prediction_each_day} shows the predictions of the SUNSET nowcast model on each of the test days compared with the ground truth.

The results show that the SUNSET nowcast model can effectively extract the information in the sky images and correlate it with the local PV panel generation. It can well approximate the sun angle equations in the sunny days with clear sky conditions and reasonably estimate the states of PV power generation under different cloudy conditions. The potential use of the nowcast model is to serve as an alternative to the traditional sensor measurement of solar irradiance or PV panel power output, which is generally expensive. A similar study by \citet{Jiang2020} has also examined the potential of using an end-to-end CNN model to estimate the state of solar irradiance.

\subsection{PV Power Generation Forecast}
The PV power generation forecast task can be mathematically described as learning a mapping ($f_F$) from historical sky image and PV power generation sequences to the future PV power generation.
\begin{equation}
f_F:(\mathcal{I},\mathcal{P})_{t_i-H:\delta:t_i} \mapsto \mathcal{P}_{t_i+T}, \mathrm{where}\  \{(\mathcal{I},\mathcal{P})_{t_i-H:\delta:t_i}, \mathcal{P}_{t_i+T}\} \in \tilde{\mathscr{D}}
\end{equation}
Here, we define forecast dataset $\tilde{\mathscr{D}}$ with $\tilde{\mathscr{D}}\subseteq {\mathscr{D}}$ and $\tilde{\mathscr{D}} = \{[(\mathcal{I},\mathcal{P})_{t_i-H:\delta:t_i}, \mathcal{P}_{t_i+T}] \mid i\in \mathbb{Z}: 1\leq i\leq M\}$, where $H$ is the length of historical terms, $\delta$ is the interval between historical terms, $T$ is the forecast horizon, and $M$ is the total number of samples in $\tilde{\mathscr{D}}$. To build the forecast dataset $\tilde{\mathscr{D}}$ for demonstration, we follow the parameters used in the baseline of \cite{Sun2019}, namely, $H=T=15$ min, $\delta=1$ min, and a sampling frequency of 2 min (i.e., $t_{i+1}-t_i = 2$) to obtain valid samples from $\mathscr{D}$, which leads to 161,928 samples in total for the forecast dataset. Following the definition by \citet{Yang2019guideline,Yang2020} to formalize the description of a solar forecasting task, ours can be described as \{$S^{1min}$, $R^{1min}$, $L^{15min}$, $U^{2min}$\}\footnote{For training and validation of the forecast model, we used a forecast submission update rate of 2 min ($U^{2min}$) to speed up the model development while retaining a good model performance based on the findings by \citet{Sun2019} [Note that the same term is called sampling frequency in the work by Sun \textit{et al}.]. For model testing, we used a forecast submission update rate of 1 min ($U^{1min}$).}, where $S$ is forecast span, $R$ forecast resolution, and $L$ is forecast lead time and $U$ is forecast submission update rate. A graphic illustration of these temporal parameters can be found in \cite{Yang2019operationalforecast}. For consistent comparisons with each other's model, we encourage the users to clearly state their forecasting setups.

\begin{figure}[h!]
\includegraphics[width=1.0\textwidth]{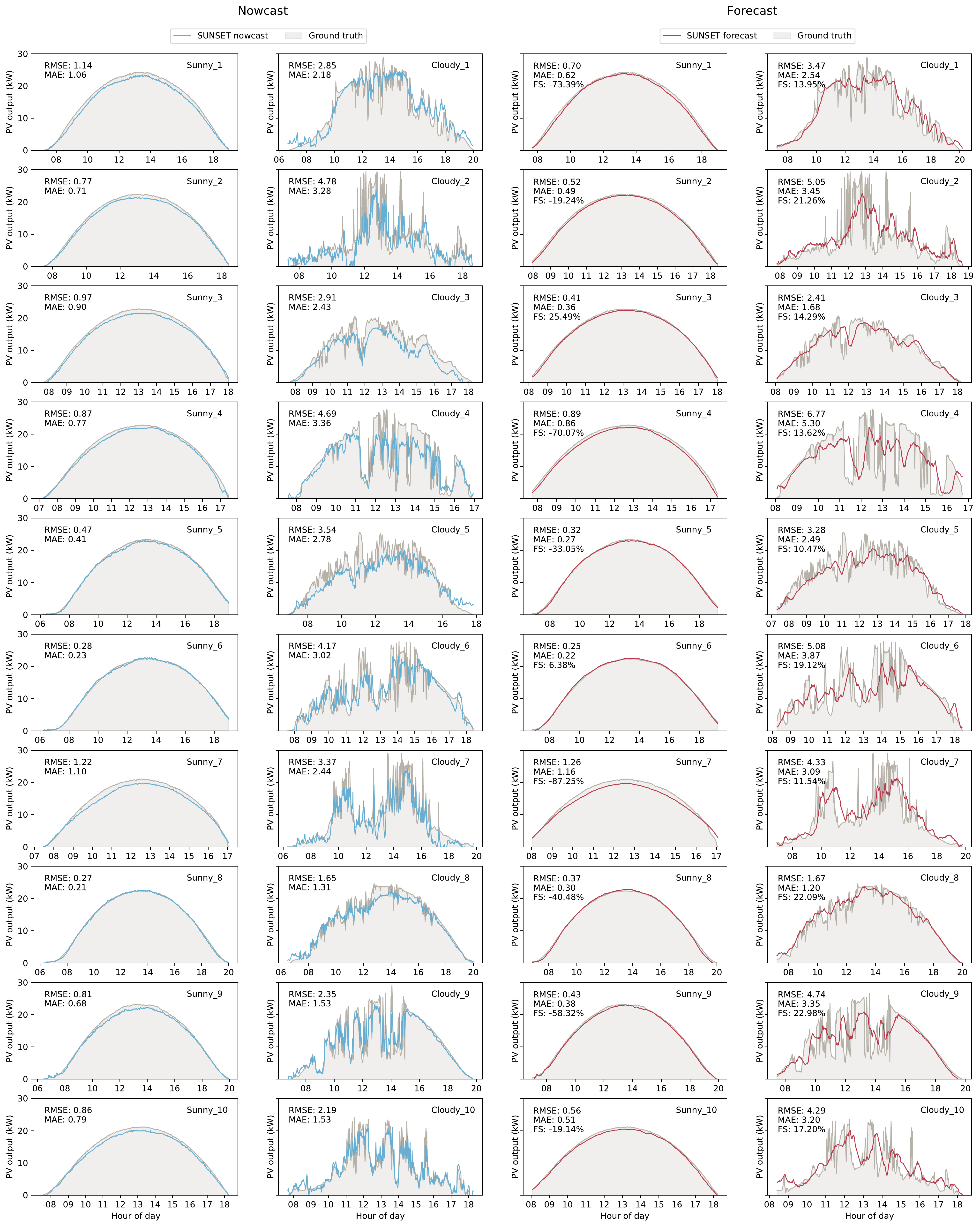}
\caption{SUNSET nowcast and forecast predictions on each of the test days. The first and second columns represent the nowcast prediction, and the third and fourth columns represent the forecast prediction. The test days’ indices and the error metrics are shown on the right and left of each panel respectively.}
\label{fig:sunset_prediction_each_day}
\end{figure}

The forecast dataset $\tilde{\mathscr{D}}$ is further separated into model development set $\tilde{\mathscr{D}}_d$ and test set $\tilde{\mathscr{D}_t}$ based on the same partition as the benchmark development set $\mathscr{D}_d$ and test sets $\mathscr{D}_t$. The processing codes we used are also open-sourced (see Appendix \ref{appendix:sample_code}) and users can either modify these parameters in our processing code based on their model input and output configurations or develop their own processing codes to build the forecast dataset. We trained the forecast model with 10-fold cross-validation, and during the test phase, the prediction is represented by the ensemble
mean of the predictions from the 10 submodels. Table \ref{tab:performance_metrics} shows the performance of the SUNSET forecast model on the test set evaluated by RMSE, MAE, and FS. Figure \ref{fig:sunset_prediction_each_day} shows the predictions of the SUNSET forecast model on each of the test days compared with the ground truth.

\begin{table}[h!]
\begin{center}
\begin{small}
\caption{The performance of SUNSET nowcast and forecast models evaluated by common error metrics}
\label{tab:performance_metrics}
\begin{tabular}{p{0.16\linewidth}p{0.12\linewidth}p{0.12\linewidth}p{0.12\linewidth}p{0.15\linewidth}}
\hline \hline
\noalign{\vskip 1mm}
Model & Test Set & RMSE (kW) & MAE (kW) & Forecast Skill (\%)\\
\noalign{\vskip 1mm}
\hline
\noalign{\vskip 1mm}
 & Sunny days & 0.80 & 0.66 & -- \\
\noalign{\vskip 1mm}
SUNSET Nowcast & Cloudy days & 3.34 & 2.34 & -- \\
\noalign{\vskip 1mm}
 & Overall & 2.43 & 1.50 & -- \\
\noalign{\vskip 1mm} \hline \noalign{\vskip 1mm}
  & Sunny days & 0.61 & 0.50 & - 45.80 \\
\noalign{\vskip 1mm}
SUNSET Forecast & Cloudy days & 4.27 & 2.95 & 17.03\\
\noalign{\vskip 1mm}
  & Overall & 3.03 & 1.71 & 16.44\\
\noalign{\vskip 1mm}
\hline \hline
\end{tabular}
\end{small}
\end{center}
\end{table}

The results show that the SUNSET forecast model can well predict the 15-min ahead future power generation on the sunny days but it struggles somehow on the cloudy days. In cloudy conditions, SUNSET tends to avoid large fluctuations by predicting conservatively towards the average. It helps the model in capturing the general trend, however, it usually fails in predicting the right magnitudes of the peaks and dips, which leads to one of the major prediction errors. Another type of error is associated with temporal lags in prediction. The model has a larger tendency of following the trend of the past observations than actively anticipating future events. These two aspects of errors direct to the following future research directions: (1) generating probabilistic prediction rather than point prediction to deal with the uncertainty in the sudden changes of PV power generation; (2) developing models for better prediction of the cloud movement to deal with the systematic temporal lags in forecasting. Compared with the persistence model, SUNSET outperforms it by 16.4\% overall in FS on all of the test days and by 17\% on cloudy days, although it under-performs the persistence model by 46\% on sunny days. To this end, using different types of models for different sky conditions could boost the accuracy of the forecast \cite{Nie2020}.

\section{Conclusion}
We introduced a curated dataset named SKIPP'D with the goal of providing a standardized benchmark for the solar forecasting community to evaluate and compare different solar forecasting models as well as to facilitate the research of image-based solar forecasting using deep learning and other related areas such as sky image segmentation, cloud type classification and cloud movement forecasting. Two levels of data are provided in this dataset including a processed benchmark dataset containing 1-min down-sampled sky images and PV output pairs and a raw dataset containing high-resolution sky videos, images and PV power generation data. We have also detailed the data processing steps used to obtain the benchmark dataset and demonstrated two use cases based on the benchmark dataset for users' reference and future studies.

\section{Acknowledgement}
The authors thank Jacques de Chalendar from Stanford University who helps us access the PV power generation history data. The authors would also like to acknowledge the Stanford Research Computing Center for providing the computational resources to conduct the experiments in this study. The authors are also grateful to Amy Hodge from Science and Engineering Resource Group at Stanford Libraries for facilitating the datasets depositing.

\begin{appendices}
\section{Data repository}
\label{appendix:data_repo}
The benchmark data is available at \url{https://purl.stanford.edu/dj417rh1007} and the raw data is deposit by year given its large size. The 2017 raw data is available at \url{https://purl.stanford.edu/sm043zf7254} and the links to 2018 and 2019 data can be found in the "Related items" elsewhere on the same web page. All datasets are licensed under CC-BY-4.0. Table \ref{tab:list_avai_files} shows the list of all data files included in this dataset. Although every effort was made to ensure the quality of the data, no guarantees are implied by the authors of the dataset.

\begin{table}[h!]
\begin{center}
\begin{small}
\caption{List of available files in the data repository}
\label{tab:list_avai_files}
\begin{tabular}{>{\raggedright}p{0.35\linewidth}>{\raggedright}p{0.13\linewidth}>{\raggedright}p{0.12\linewidth}>{\raggedright\arraybackslash}p{0.30\linewidth}}
\hline\hline
\noalign{\vskip 1mm}
File & Type & Size & Description \\
\noalign{\vskip 1mm}
\hline
\noalign{\vskip 1mm}
\texttt{2017\_2019\_images\_pv\_processed.hdf5} & Benchmark data & 4.16 GB & A file-directory like structure consisting of two groups: \texttt{"trainval"} and \texttt{"test"}, for storing model development set and test set, respectively, with each group containing two datasets: \texttt{"images\_log"} and \texttt{"pv\_log"}, which stores the processed images stored as 8-bit RGB (256 color levels) data and PV generation data in the unit of kW from all three years stored in Python NumPy array format. \\
\noalign{\vskip 1mm}
\texttt{times\_trainval.npy} & Benchmark data & 10.8 MB & Python NumPy array of time stamps stored in Python \textit{datetime} format corresponding to development set in \textit{.hdf5} file. \\
\noalign{\vskip 1mm}
\texttt{times\_test.npy} & Benchmark data & 434 KB & Python NumPy array of time stamps stored in Python \textit{datetime} format corresponding to test set in \textit{.hdf5} file. \\ 
\noalign{\vskip 1mm} \hline \noalign{\vskip 1mm}
\texttt{\{Year\}\_\{Month\}\_videos.tar} & Raw data & 2017: 500 GB; 2018: 640 GB; 2019: 449 GB & Tar archives with daytime 2048 $\times$ 2048 sky videos (\textit{.mp4}) recorded at 20 frames per second for each month from 2017/03 to 2019/12. \\
\noalign{\vskip 1mm}
\texttt{\{Year\}\_\{Month\}\_images\_raw.tar} & Raw data & 2017: 28.2 GB; 2018: 50.1 GB; 2019: 55.3 GB & Tar archives with daytime 2048 $\times$ 2048 sky images (\textit{.jpg}) captured at 1-min intervals for each month from 2017/03 to 2019/12.\\
\noalign{\vskip 1mm}
\texttt{\{Year\}\_pv\_raw.csv} & Raw data & 2017: 16.7 MB; 2018: 16.6 MB; 2019: 13.7 MB & One-min PV generation data for the year 2017, 2018 and 2019. The unit of PV generation data is kW.\\
\noalign{\vskip 1mm}
\hline\hline
\end{tabular}
\end{small}
\end{center}
\end{table}


\section{Code Base}
\label{appendix:sample_code}
As part of the data release, we include the data processing and baseline model implementation code written in Python 3.6, which can be found on the Github Repository \url{https://github.com/yuhao-nie/Stanford-solar-forecasting-dataset}.  All code files are licensed under the MIT license. Table \ref{tab:list_code_files} shows the list of all the code files included along with this dataset. Users can either use the reference codes we provided here or customize their own data processing pipeline.

\begin{table}[h!]
\begin{center}
\begin{small}
\caption{List of available code files}
\label{tab:list_code_files}
\begin{tabular}{>{\raggedright}p{0.35\linewidth}>{\raggedright}p{0.18\linewidth}>{\raggedright\arraybackslash}p{0.36\linewidth}}
\hline\hline
\noalign{\vskip 1mm}
File & Type & Description \\
\noalign{\vskip 1mm}
\hline
\noalign{\vskip 1mm}
\texttt{data\_preprocess\_snapshot\_only.ipynb} & Data processing code & Jupyter Notebook used to capture images from the video stream at designated frequency. \\
\noalign{\vskip 1mm}
\texttt{data\_preprocess\_pv.ipynb} & Data processing code & Jupyter Notebook used to process the raw PV power generation history. \\
\noalign{\vskip 1mm}
\texttt{data\_preprocess\_nowcast.ipynb} & Data processing code & Jupyter Notebook used to down-sample the image frames, filter out the invalid frames and match images with the concurrent PV data, and partition model development and testing sets. \\
\noalign{\vskip 1mm}
\texttt{data\_preprocess\_forecast.ipynb} & Data processing code & Jupyter Notebook used to generate valid samples for the forecast task. \\
\noalign{\vskip 1mm} \hline \noalign{\vskip 1mm}
\texttt{SUNSET\_nowcast.ipynb} & Model code & Jupyter Notebook used to implement the SUNSET nowcast model to correlate PV output to contemporaneous images of the sky, including model training, validation and testing. \\
\noalign{\vskip 1mm}
\texttt{SUNSET\_forecast.ipynb} & Model code & Jupyter Notebook used to implement the SUNSET forecast model to predict 15-min ahead minutely-averaged PV output, including model training, validation and testing. \\
\noalign{\vskip 1mm} \hline \noalign{\vskip 1mm}
\texttt{Relative\_op\_func.py} & Helper function code & Helper functions for calculating theoretical PV power output under clear sky condition and the clear sky index. \\
\noalign{\vskip 1mm}
\hline\hline
\end{tabular}
\end{small}
\end{center}
\end{table}

\end{appendices}

\clearpage
\bibliography{references}

\end{document}